\documentclass[letterpaper, 10 pt, conference]{ieeeconf}  %

\usepackage{amsmath}
\usepackage{amssymb}  %
\usepackage{amsfonts}
\usepackage{comment}
\usepackage{xcolor}
\usepackage{mathtools}
\usepackage{xcolor}

\newtheorem{proposition}{Proposition}
\usepackage{graphicx}
 \usepackage{booktabs}

\newcommand{\R}{\mathbb{R}}
\newcommand{\N}{\mathbb{N}}

\newcommand{\card}{{\textit{card}}}
\newtheorem{theorem}{Theorem}
\newtheorem{definition}{Definition}

\usepackage[vlined, ruled, boxed, linesnumbered]{algorithm2e}
\SetKw{Continue}{continue}
\SetKw{Break}{break}
\SetKwFor{ForAll}{for all}{do}{end}
\SetKwFor{ParForAll}{for all}{in parallel with index $j$ do}{end}
\SetKwFor{ParForAllNoIndex}{for all}{in parallel do}{end}
\SetKwRepeat{Do}{do}{while}
\SetKwInOut{Parameter}{Parameters}
\SetKwComment{Comment}{$\triangleright$\ }{}

\newcommand{\ind}[1]{1\left\{ #1 \right\}}

\IEEEoverridecommandlockouts                              %

\title{\LARGE \bf Symbolic Abstractions From Data: A PAC Learning Approach }

\author{Alex Devonport$^{*}$, Adnane Saoud$^{*}$, and Murat Arcak%
\thanks{$^*$Both authors have provided equal contribution to the work in this paper.}
\thanks{
Alex Devonport, Adnane Saoud and Murat Arcak are with the Dept. of Electrical Engineering and Computer Sciences, University of California, Berkeley, \texttt{\{alex\_devonport,asaoud,arcak\}@berkeley.edu}.
}
\thanks{
This work was supported in part by the grants ONR N00014-18-1-2209, AFOSR 
FA9550-18-1-0253, NSF ECCS-1906164.
}%
}

\begin{document}

\maketitle
\thispagestyle{empty}
\pagestyle{empty}

\begin{abstract}
\begin{comment}
Symbolic control techniques have recently shown a great success at controller synthesis for dynamical systems under complex logic specifications. A critical step in symbolic control is the construction of a symbolic (discrete) abstraction, a finite-state system whose behaviour mimics that of a given continuous-state system. These methods used to compute symbolic abstractions require knowledge of an accurate closed-form model, making symbolic control techniques inaccessible to systems with unknown dynamics. To address this limitation, we present a new data-driven approach for computing guaranteed finite-state abstractions. This approach does not require closed-form dynamics, but instead only the ability to evaluate successors of individual points for given inputs. In order to provide guarantees, we use the Probably Approximately Correct (PAC) statistical framework. To incorporate PAC-style guarantees into symbolic control, we first propose a new behavioural relationship with an appropriate refinement procedure, making it possible to relate the original system with its symbolic abstraction. We then show how the symbolic abstraction can be constructed in the context of this new behavioural relationship. Moreover, for the constructed abstraction, we provide guarantees of accuracy and confidence, which can be made arbitrarily precise by appropriately choosing the number of required data. Finally, an illustrative example is provided showing the merits of the proposed approach.
\end{comment}
Symbolic control techniques aim to satisfy complex logic specifications. A critical step in these techniques is the construction of a symbolic (discrete) abstraction, a finite-state system whose behaviour mimics that of a given continuous-state system. The methods used to compute symbolic abstractions, however, require knowledge of an accurate closed-form model. To generalize them to systems with unknown dynamics, we present a new data-driven approach that does not require closed-form dynamics,  instead relying only the ability to evaluate successors of each state under given inputs. To provide guarantees for the learned abstraction, we use the Probably Approximately Correct (PAC) statistical framework. We first introduce a PAC-style behavioural relationship and an appropriate refinement procedure. We then show how the symbolic abstraction can be constructed to satisfy this new behavioural relationship. Moreover, we provide PAC bounds that dictate the number of data required to guarantee a prescribed level of accuracy and confidence. Finally, we present an illustrative example.
\end{abstract}

\section{Introduction}

Research at the interface between formal methods and control theory has given rise to symbolic control~\cite{tabuada2009verification,belta2017formal,saoud2019compositional}, which deals with control of dynamical systems with logic specifications~\cite{baier2008principles}. A key ingredient of symbolic control is a finite abstraction, i.e. a dynamical system with a finite number of states and inputs, also called symbolic model, constructed from the original system. When the concrete and abstract systems are related by a behavioral relation, such as an {\it approximate alternating simulation relation}~\cite{tabuada2009verification}, showing that the trajectories of the abstraction mimic the ones of the original system, the discrete controller synthesized for the abstraction can be refined into a hybrid controller for the original system. Finite abstractions enable the use of techniques developed in the areas of supervisory control of discrete event systems \cite{cassandras2009introduction} and algorithmic game theory \cite{bloem2012synthesis}.

The abstractions studied in the literature generally rely on a mathematical model of the system, such as a physics-based first principle model. A closed-form model, however, may not be available when dealing with cyber-physical systems consisting of components of a different nature, described by differential equations, lookup tables, transition systems and hybrid automata. In this case, a common practice is to make use of learning-based approaches.
A number of learning-based approaches have been proposed recently for control of unknown systems {\it without} using symbolic models.
These results study
particular classes of systems, such as feedback-linearizable systems  in~\cite{tabuada2017data}, linear-time invariant systems in~\cite{de2019formulas,berberich2020data,coulson2019data}, and address particular types of specifications, such as safety in~\cite{fisac2018general}, stability  in~\cite{berkenkamp2017safe,de2019formulas,tabuada2017data,berberich2020data, devonport2020bayesian}, and trajectory tracking in~\cite{coulson2019data}. 
Learning a symbolic abstraction offers several benefits over the aforementioned results. First, our approach is universal and makes it possible to deal with general nonlinear systems subject to constraints. Second, 
we can deal with complex specifications, such as those expressed  in linear temporal logic over finite traces~\cite{de2013linear}. 

In this paper, we propose a data-driven approach to construct system abstractions whose fidelity to the continuous system is guaranteed by a PAC bound~\cite{valiant1984theory}. In contrast to classical asymptotic statistical guarantees, which only guarantee certain behaviours in the limiting case of infinite data, PAC bounds %
restrict
the probability of error that a statistical estimate makes with finite data. Additionally, PAC bounds are distribution-free, meaning that the guarantee holds regardless of the specific probabilistic nature of the problem. This makes PAC bounds a suitable type of guarantee for statistical approaches to control theory, where random variables are propagated through unknown functions (e.g. the state transition function), and guarantees must be made with finite data. Indeed, PAC analysis is already a popular tool for data-driven methods in safety-critical control, being used for such applications as reachability analysis~\cite{devonport2020data,devonport2020estimating}, model-predictive control~\cite{hewing2019scenario}, and safety verification of autonomous vehicles~\cite{fan2017dryvr,qi2018dryvr}. However, PAC analysis has not, to the best of our knowledge, been applied to the construction of finite abstractions for symbolic control.

Classical approaches to the construction of model-based symbolic abstractions rely on reachability analysis. In this paper, the construction of data-driven abstractions is based on the PAC approach, where the discrete successors are computed as solutions to a statistical problem of empirical risk minimization.
The contribution of the paper is twofold: First, we generalize the classical concept of alternating simulation relation~\cite{tabuada2009verification} to incorporate the types of guarantees made by PAC learning, together with an appropriate refinement procedure. Second, for a given dynamical system with unknown dynamics, we present a new approach to the construction of data-driven symbolic abstractions that are related to the original dynamical system by a PAC approximate alternating relation. Moreover, we show that the provided statistical guarantees of the constructed symbolic abstraction, in terms of accuracy and confidence, can be made arbitrarily small  by an appropriate choice of the number of data.

Related results in the literature include~\cite{hashimoto2020learning}, which studies discrete-time nonlinear control systems with unmodeled dynamics, represented as a Gaussian Process.
A symbolic model is constructed using
Lipschitz continuity bounds of the unmodeled dynamics, learned from training data. Another related reference is~\cite{lavaei2020formal}, which constructs abstractions for stochastic discrete time nonlinear systems and co-safe LTL specifications. Again, a Lipschitz continuity bound is learned from data and used to build a model, which  is then used to construct the symbolic abstraction. In this paper, we learn the abstraction directly from data without resorting to the model, which enables us to construct the abstraction for larger classes of systems without imposing any smoothness or regularity assumptions on the model. 

\begin{comment}
In the spirit, the closest works in the literature are~\cite{hashimoto2020learning,lavaei2020formal}. In~\cite{hashimoto2020learning}, the authors consider partially known discrete-time nonlinear control systems. Under the assumption that the unmodeled part of the dynamics follows a Gaussian Process, they estimate the Lipschitz continuity bounds of the unmodeled dynamics from training data. Then, they use the learned Lipschitz bound to construct a symbolic model. In~\cite{lavaei2020formal}, the authors propose an approach to construct abstractions for stochastic discrete time nonlinear systems and co-safe LTL specifications, under the assumption of the knowledge of the global Lipschitz continuity bound of the model. In the aforementioned approaches, a Lipschitz continuity bound is learned from data and used to construct a model, the latter is then used to construct the symbolic abstraction. In this paper, we learn the abstraction directly from data without resorting to the model, which makes it possible to construct the abstraction for larger classes of systems without imposing any smoothness or regularity assumptions on the model.%
\end{comment}

The remainder of this paper is organized as follows. In Section~\ref{sec:2} we recall the classical approach to the construction of symbolic abstractions for the considered class of systems and the statistical learning tools used in the paper. In Section~\ref{sec:3} we introduce the concept of PAC-approximate alternating simulation and show how this new concept makes it possible to learn the symbolic abstraction from data, while providing statistical guarantees. Finally, Section~\ref{sec:4} presents numerical results highlighting the merits of the proposed approach.

\textbf{Notations:} The symbols $ \mathbb{N}$, $ \mathbb{N}_{>0} $, and $\R$ and $\R_{>0}$ denote the set of positive integers, non-negative integers, real and non-negative real numbers, respectively. Given a relation $\mathcal{R} \subseteq X_1 \times X_2$ and  $x_1 \in X_1$ we define $\mathcal{R}(x_1)=\{x_2 \in X_2 \mid (x_1,x_2) \in \mathcal{R}\}$. Similarly, for $x_2 \in X_2$ we define $\mathcal{R}^{-1}(x_2)=\{x_1 \in X_1 \mid (x_1,x_2) \in \mathcal{R}\}$. For a set $A$, $\card(A)$ denotes the cardinality of $A$. A relation $\mathcal{R} \subseteq X_1 \times X_2$ is said to be deterministic if for any $x_1 \in X_1$, $\card(\mathcal{R}(x_1)) \leq 1$. Given $N \in \mathbb{N}_{>0}$ and a metric space $Y$ equipped with a metric $\mathbf{d}$, $Y^{\leq N}$ denotes the set of finite sequences of elements of $Y$ with length bounded by $N$. For a set $C \subseteq Y^{\leq N}$ and $\varepsilon \geq 0$, $\mathcal{B}_{\varepsilon}(C) \subseteq Y^{\leq N}$ is defined by $\mathcal{B}_{\varepsilon}(C)=\{z_0,z_1,\ldots,z_m \in Y^{\leq N} \mid \exists~  y_0,y_1,\ldots,y_m \in C \text{ with } \mathbf{d}(z_i,y_i) \leq \varepsilon,\; \forall i \in \{1,\ldots,m\}\}$.
Given sets $A_1,\dotsc,A_n\subseteq X$, let $\bigtimes_{i=1}^n A_i=\{(x_1,\dotsc,x_n):x_i\in A_i, i=1,\dotsc,n\}$ denote their Cartesian product.

\section{Symbolic abstractions for dynamical systems}
\label{sec:2}

In this section, we first define the class of systems considered in the paper. We then recall the classical approach to construct their symbolic abstractions.

\subsection{Preliminaries}

First, we review the notion of a \textit{transition system}~\cite{tabuada2009verification}. This notion allows us to describe the concrete dynamical system and its symbolic abstraction in a unified framework.
\begin{definition}\label{definition6}
	A transition system is a tuple $S=(X,X^0,U,\Delta,Y,H)$, where $ X $ is the set of states, $X^0 \subseteq X$ is the set of initial states, $U$ is the set of inputs, $ \Delta \subseteq X\times U\times X $ is the transition relation, $Y$ is the set of outputs and $H$ is the output map. When $ (x,u,x')\in\Delta$, we use the alternative representation  $ x'\in \Delta(x,u),$  where state $ x' $ is called a $ u $-successor (or simply successor) of state $ x $ under input $ u\in U $. 
\end{definition}

The transition system is said to be:
\begin{itemize}
	\item \textit{finite} (or \textit{symbolic}), if sets $ X $ and $ U $ are finite.
	\item \textit{deterministic}, if there exists at most one $ u $-successor of $ x $, for any $ x\in X $ and $ u\in U $.  
\end{itemize}

Given $x\in X$, the set of enabled (admissible) inputs for $x$ is denoted by $U^a(x)$ and defined as $U^a(x)=\{u\in U \mid \Delta(x,u)\neq \emptyset\}$. A trajectory of the transition system $S$ is a finite or infinite sequence of transitions $\sigma=(x_0,u_0),(x_1,u_1)(x_2,u_2)\ldots$ where $x_{i+1} \in \Delta(x_i,u_i)$ for $i \in \mathbb{N}$. The output behavior associated with the trajectory $\sigma$ is the sequence of outputs $\sigma_y=y_0,y_1,y_2,\ldots$ where $y_i=H(x_i)$ for all $i \in \mathbb{N}$. In this paper, we are interested in finite-time specifications. Given a constant $N \in \mathbb{N}_{>0}$, a specification $\phi \subseteq Y^{\leq N}$ for the transition system $S$ encodes a set of finite-time desirable behaviors for which the length of the output behaviour is bounded by $N$.

We now define the concept of control policy for transition systems.
\begin{definition}
   A control policy for the transition system
   $S=(X,X^0,U,\Delta,Y,H)$
   is a pair $(X^0_{\mathcal{C}},\mathcal{C})$ consisting of:
    \begin{itemize}
        \item A set of initial states $X^0_{\mathcal{C}}$;
        \item A control law $\mathcal{C}:X^{\leq N} \rightarrow U$ that takes a finite state sequence $x_0x_1\ldots x_k$ and outputs a control $u_k$ for all time $k\in \mathbb{N}$.
    \end{itemize}
\end{definition}

\medskip

When the control policy is placed in closed loop with the transition system $S$, a controlled output behaviour $\sigma_y=y_0,y_1,y_2,\ldots ,y_{m-1}$ satisfies: $x_0 \in X^0_{\mathcal{C}}$, $x_{k+1} \in \Delta(x_k,u_k)$ with $u_k=\mathcal{C}(x_0x_1\ldots x_k)$ and $y_k=H(x_k)$ for all $k \in \{0,1,\ldots,m-1\}$, $m \leq N$. We use $\mathcal{B}_S(X^0_{\mathcal{C}},\mathcal{C}) \subseteq Y^{\leq N}$ to denote the set of all possible output behaviours when the control policy is placed in closed loop with the transition system $S$.

In the sequel, we consider the approximate relationship for transition systems based on the notion of alternating simulation relation to relate abstractions to concrete systems. 

\begin{definition}
	\label{Def:altsimu}
	Let $S_1=(X_1,X_1^0,U_1,\Delta_1,Y_1,H_1)$ and $S_2=(X_2,X_2^0,U_2,\Delta_2,Y_2,H_2)$ be two transition systems such that $Y_1$ and $Y_2$ are subsets of the same metric space $Y$ equipped with a metric $\mathbf{d}$. For $\varepsilon \geq 0$, a deterministic relation $\mathcal R \subseteq X_1\times X_2$ is said to be an $\varepsilon$-approximate alternating simulation relation from $S_2$ to $S_1$, if it satisfies:
	\begin{itemize}
		\item[(i)] $\forall x_2^0\in X_2^0$, $\exists  x_1^0\in X_1^0$ such that $(x_1^0,x_2^0)\in \mathcal{R}$;
		\item[(ii)]   $\forall (x_1,x_2)\in \mathcal R$, $\mathbf{d}(H_1(x_1),H_2(x_2))\leq \varepsilon$;
		\item[(iii)]  $\forall (x_1,x_2)\in \mathcal R$, $\forall u_2 \in U^a_2(x_2)$,
		$\exists u_1 \in U^a_1(x_1)$ such that $ \forall x_1'\in  \Delta_1(x_1,u_1)$, $\exists x_2'\in  \Delta_2(x_2,u_2)$ satisfying $(x_1',x_2')\in \mathcal R$.
	\end{itemize}
\end{definition}

\medskip

We denote the existence of an $\varepsilon$-approximate alternating simulation relation from $S_2$ to $S_1$ by $ S_2\preccurlyeq_{\varepsilon} S_1$.

\begin{comment}
\subsection{Discrete-time control systems}
\label{sec:discrete}

\begin{definition}
\label{def:dtsystem}
	A discrete-time control system $\Sigma$ is a tuple $\Sigma=(X,X^0,U,f)$, consisting of a set of states $X$ and a set of control inputs $U$. The function $f:X\times U \longrightarrow X$ represents the dynamics of the system $\Sigma$ and is called the transition function. 
\end{definition}
The evolution of the state of the discrete-time control system $\Sigma$ is described as
\begin{equation}
\label{dis_sys}
x(k+1) = f(x(k),u(k)), ~x(0) \in X^0,
\end{equation}
where $x(k)\in X\subset\mathbb{R}^n$ is a state and $u(k) \in U\subset\mathbb{R}^m$ is a control input. In this paper, we assume that the sets $X$ and $U$ are bounded. 

The control system $\Sigma=(X,X^0,U,f)$ can be represented as a transition system $S=(X,X^0,U,\Delta,Y,H)$ where the set of states $X$, the set of initial states $X^0$, and the set of inputs $U$ are inherited from the discrete-time control system $\Sigma$. The transition relation $\Delta$ corresponds to the transition function $f$. The output set is $Y=X$ and the output map is the identity map defined for $x \in X$ as $H(x)=x$.

\end{comment}

\subsection{Symbolic abstraction}
\label{subsec:symbolic_abstraction}

In this paper, we model a dynamical system $S=(X,X^0,U,\Delta,Y,H)$ as a transition systems (see Definition~\ref{definition6}), where $X \subseteq \mathbb{R}^n$ is the set of states, $X^0 \subseteq X$ is the set of initial states, $U \subseteq \R^p$ is the set of inputs, $\Delta:X\times U \rightarrow X$ is a deterministic transition relation, $Y=X$ is the set of outputs and $H$ is the output map satisfying $H(x)=x$ for all $x \in X$. The dynamical system $S$ is generally equipped with a continuous state and input spaces $X$ and $U$, respectively, and has an infinity of possible transitions. The objective of this part is to recall the classical approach to construct a symbolic (finite or discrete) abstraction $S_d=(X_d,X_d^0,U_d,\Delta_d,Y_d,H_d)$ for the concrete system $S$. 

\subsubsection{Discretization}
To construct the state and input sets $X_d$ and $U_d$ for the symbolic abstraction $S_d$, we rely on the discretization of the continuous state-space $X$ and input sets $U$. First, we approximate the set of inputs $U$ with a finite number of values $n_u$:
$U_d=\big\{\mathsf{u}_{\ell}\in U|\; \ell=0,\dots,n_u-1 \big\}$.
We discretize the state-space, using a discretization precision $\eta>0$, into $n_x\ge 1$ states using a finite partition $X_d$ of the set $X$. Each element $q$ of the partition is a half-closed interval $q=(\underline{x}^q,\overline{x}^q]$. We also define the quantizer $Q_{X_d}: X \rightarrow X_d$ associating each continuous state to its symbolic counterpart: for $x \in X$ and $q \in X_d$, $Q_{X_d}(x)=q$ if and only if $x\in q$. The set of discrete initial states is defined as $X_d^0=X_d \cap X^0$. The discrete output set is $Y_d=X$ and the output map is defined for $q=(\underline{x}^q,\overline{x}^q] \in X_d$ as $H_d(q)=\frac{\underline{x}^q+\overline{x}^q}{2}$.

\subsubsection{Transition relation} 
The transition relation $\Delta_d \subseteq X_d\times U_d \times X_d$ abstracting the continuous dynamics can be defined as follows: for $q,q'\in X_d$, $u\in U_d$, $q' \in \Delta_d(q,u)$ if and only if $\Delta(q,u) \in (x_1^{q'},x_2^{q'}]$, where
\begin{equation}
\label{eqn:reach}
    \Delta(q,u)=\{z \mid z=\Delta(x,u), x \in q\}
\end{equation}

Using such a construction for the symbolic abstraction, one can show that the abstraction $S_d$ is approximately alternatingly simulated by the concrete system $S$, i.e, $S_d \preccurlyeq_{\eta} S$, where $\eta>0$ is the state-space discretization parameter used to construct the symbolic abstraction~\cite{tabuada2009verification}.

In model-based approaches the key ingredient for finding the discrete successors is the computation of reachable sets. In this paper, the construction of the abstraction is based on the PAC approach, where the discrete successors are computed as solutions to a statistical problem of \emph{empirical risk minimization} which satisfies a PAC bound. 

\subsection{Empirical Risk Minimization and PAC Bounds}
\label{subsec:empirical_risk_minimization_and_pac_bounds}

An empirical risk minimization problem comprises a class $\mathcal{C}$ of sets $c\in X$, called the concept class, a \emph{loss function} $\ell:(\mathcal{C}\times X)\to \R$, and a set of sample points $x^{(1)},\dotsc,x^{(M)}$ that are independent and identically distributed according to an unknown probability measure $P$. These components define the risk function $r(c)=\int_X \ell(c,x)dP(x)$. The goal is to find the concept $c\in\mathcal{C}$ that minimizes $r(c)$, but this is not possible since $P$ is unknown. We instead find the concept that minimizes the \emph{empirical risk} $\hat{r}(c)=\frac{1}{M} \sum_{i=1}^M \ell(c,x^{(i)})$. 

Under certain conditions on the concept class\footnote{One common condition on the concept class is that its \emph{Vapnik-Chervonenkis (VC) dimension}, a combinatorial measure of complexity, be finite~\cite{blumer1989learnability}. All concept classes considered in this paper have finite VC dimension by construction.}, the difference between $r(c)$ and $\hat{r}(c)$ can be uniformly bounded with high probability by any constant $\mu > 0$ for a sufficiently large (but finite) sample size $M$. In the case of concept classes of finite cardinality, we have the following bound:

\begin{proposition}[\cite{blumer1989learnability}, Theorem 2.2]
\label{prop:finite_concept_pac}
Consider the confidence parameter $\delta \in (0,1)$ and the accuracy parameter $\mu \in(0,1)$. Let $r$ be the risk function and $\hat{r}$ the empirical risk function, defined above. If the sample size is chosen such that $M \ge \frac{1}{\mu}\log\frac{\text{card}(\mathcal{C})}{\delta}$, then the following holds:
\begin{equation}
    \label{eq:pac_def}
    P^{M}\left(\sup_{c\in\mathcal{C}} |r(c)-\hat{r}(c)| \le \mu\right) \ge 1-\delta,
\end{equation}
where $P^{M}$ denotes the product measure\footnote{The product of measures $P_i$, $i=1,\dotsc,n$, is defined as a measure $P_{\text{prod}}$ which satisfies $P_{\text{prod}}(\bigtimes_{i=1}^n A_i)=\prod_{i=1}^nP_i(A_i)$ when $A_i$ is $P_i$-measurable for all $i$. If the $P_i$ are all probability measures, which is always the case in this paper, then the product measure is unique.} of $M$ copies of $P$.
\end{proposition}

\bigskip

Bounds of the form~\eqref{eq:pac_def} are called \emph{PAC bounds} after the Probably Approximately Correct (PAC) framework of Valiant~\cite{valiant1984theory}. Since the product measure $P^M$ is the probability measure of the random vector $(x^{(1)},\dotsc,x^{(M)})$, equation ~\eqref{eq:pac_def} asserts that the probability of choosing samples $(x^{(1)},\dotsc,x^{(M)})$ that yield, for all concepts, an empirical risk that is within $\mu$ of the actual risk, is at least $1-\delta$. This ensures that the empirical risk is an acceptable proxy for the actual risk, satisfying a quantitative error bound with high probability.

We consider in particular the \emph{zero-one membership loss}  $\ell(c,x)=\ind{x\notin c}$. For this loss function, the associated risk is $r(c)=\int_X \ell(c,x)dP(x) = 1-P(c)$, the probability mass of the complement of $c$ under the measure $P$, and the PAC bound~\eqref{eq:pac_def} becomes the double probability inequality
\begin{equation}
    \label{eq:pac_prob}
    P^{M}\left(\sup_{c\in\mathcal{C}} |P(c)-\hat{P}(c)| \le \mu\right) \ge 1-\delta,
\end{equation}
which assures with confidence $1-\delta$ that $\hat{P}(c)$ estimates $P(c)$ with an error bounded by $\mu$ for all $c$.

\section{Main result}
\label{sec:3}

\subsection{Problem formulation and standing assumptions}

The goal of this paper is to construct a PAC-guaranteed symbolic abstraction of the concrete dynamical system $S=(X,X^0,U,\Delta,Y,H)$ for the case where the transition relation $\Delta$ is unknown. In particular, we propose here a statistical learning-based approach, in which the symbolic model is learned from 
data. Moreover, we will provide a constructive procedure to collect these data.

To provide the statistical structure to the problem, we use the probability measure space $(X, \mathcal{B}(X), \mathbb{P})$, where $X$ denotes the state space of the concrete system, $\mathcal{B}(X)$ denotes the Borel sets of $X$, and $\mathbb{P}$ denotes a probability measure on $\mathcal{B}(X)$.\footnote{We do not require that the measure $\mathbb{P}$ be explicitly defined: it serves to formally define the random variables in the sequel, but explicit knowledge of $\mathbb{P}$ is not necessary for calculations.}

For the statistical approach, we require that the unknown transition relation $\Delta$ be a deterministic\footnote{Since the transition system $S$ corresponds to the concrete dynamical system, the determinism assumption is generally satisfied.} function that is measurable with respect to $\mathcal{B}(X)$ for fixed $u\in U$, so that the image of a random variable on $X$ under $\Delta$ with fixed $u$, i.e $\Delta(\cdot,u)$, is itself a well-defined random variable on $X$. 
We also assume that we may make any finite number of evaluations $\Delta(x,u)$ for $x\in X$, $u\in U$. Essentially, the transition relation is treated as a 
\emph{black-box model}: while we assume no direct knowledge of $\Delta$ (other than determinism and measurability), we assume that the system is available for collecting data. 
The measurability assumption required in this paper is less restrictive than the global Lipschitz assumption on the unknown transition function $\Delta$, which is stipulated in~\cite{hashimoto2020learning} for the construction of symbolic abstractions for (partially) unknown dynamical systems.

\subsection{PAC behavioural relationship}

To incorporate the types of guarantees made by PAC learning into the refinement procedure, we require a generalized alternating relation which admits PAC bounds. In this section we construct such a generalization, which recovers the classical alternating simulation relation as a limiting case.

Consider the transition systems $S_1=(X_1,X_1^0,U_1,\Delta_1,Y_1,H_1)$, $S_2=(X_2,X_2^0,U_2,\Delta_2,Y_2,H_2)$ and a relation $\mathcal{R} \subseteq X_1 \times X_2$, assume that the transition system $S_1$ is deterministic and that $U_2 \subseteq U_1$. For each $x_2\in X_2$, let $z_{x_2}$ denote a random variable supported on $\mathcal{R}^{-1}(x_2)\subseteq X_1$. For instance, if $\mathcal{R}^{-1}(x_2)$ is compact, then we may choose $z_{x_2}$ to be the uniform random variable on $\mathcal{R}^{-1}(x_2)$.
	Given $z_{x_2}$, let $z_{x_2}^{(1)},\dotsc,z_{x_2}^{(M)}$ denote a set of $M$ independent and identically distributed samples drawn from $z_{x_2}$. Note that $(z_{x_2}^{(1)},\dotsc,z_{x_2}^{(M)})$ is itself a random variable, whose probability measure $P_{z_{x_2}^{(1)},\dotsc,z_{x_2}^{(M)}}$ is simply the product measure of $M$ copies of $P_{z_{x_2}}$. For a fixed $u_2\in U_2 \subseteq U_1$, let $\Delta_{1,x_2,u_2}=\Delta_1(z_{x_2},u_2)$ denote the image of $z_{x_2}$ under $\Delta_1(\cdot, u_2)$, and let $P_{\Delta_1(z_{x_2},u_2)}$ denote the probability measure corresponding to $\Delta_1(z_{x_2},u_2)$.
	In order to define a PAC-approximate alternating simulation relation, we choose a measure $P_{z_{x_2}}$ and a corresponding sample set $z_{x_2}^{(1)},\dotsc,z_{x_2}^{(M)}$ for each $x_2\in X_2$.
	
	We have now all the ingredients to define the concept of PAC-approximate alternating simulation relation.
\begin{definition}
\label{Def:PACaltsimu}
    Let $S_1=(X_1,X_1^0,U_1,\Delta_1,Y_1,H_1)$ be a deterministic transition system and $S_2=(X_2,X_2^0,U_2,\Delta_2,Y_2,H_2)$ be a finite transition system such that $U_2 \subseteq U_1$, and $Y_1$, $Y_2$ are subsets of the same metric space $Y$, equipped with a metric $\mathbf{d}$. Let $n_x=\card(X_2)$ and $n_u=\card(U_2)$. For a precision $\varepsilon \geq 0$, a confidence parameter $\delta \in (0,1)$, and an accuracy parameter $\mu \in(0,1)$, a deterministic relation $\mathcal R \subseteq X_1\times X_2$ is said to be a $(\varepsilon,\delta,\mu)$-PAC approximate alternating simulation relation from $S_2$ to $S_1$, if it satisfies:
	\begin{itemize}
		\item[(i)] $\forall x_2^0\in X_2^0$, $\exists  x_1^0\in X_1^0$ such that $(x_1^0,x_2^0)\in \mathcal{R}$;
		\item[(ii)]   $\forall (x_1,x_2)\in \mathcal R$, $\mathbf{d}(H_1(x_1),H_2(x_2))\leq \varepsilon$;
		\item[(iii)]  $\forall (x_1,x_2)\in \mathcal R$, $\forall u_2 \in U^a_2(x_2)$,
		$\exists u_1 \in U^a_1(x_1)$ satisfying
		\begin{align}
		   &P_{\Delta_1(z_{x_2}^{(1)}, u_1),
		   \dotsc,
		   \Delta_1(z_{x_2}^{(M)}, u_1)
		   }\Big(P_{\Delta_{1,x_2,u}}\Big( x_1'=  \Delta_1(x_1,u_1):\; \nonumber\\&\quad \exists x_2'\in  \Delta_2(x_2,u_2) \text{ s.t } (x_1',x_2')\in \mathcal R\Big) \geq 1-\mu \Big) \nonumber\\  \label{eqn:PACAS} &\quad \geq 1-\delta/n_x n_u
		\end{align}
	\end{itemize}
\end{definition}
We denote the 
$(\varepsilon,\delta,\mu)$-PAC approximate alternating simulation relation from $S_2$ to $S_1$ by $ S_2\preccurlyeq^{PAC}_{(\varepsilon,\delta,\mu)} S_1$.

In this definition, $S_1$ represents the concrete system and $S_2$ represents the abstract one. While conditions (i) and (ii) are the same as in the classical approximate alternating simulation relation in Definition~\ref{Def:altsimu}, condition (iii) generalizes the one in Definition~\ref{Def:altsimu}:
it states
that for each pair of concrete and discrete states $x_1$ and $x_2$ that are in relation and for each enabled discrete input $u_2$, there exists an enabled concrete input $u_1$ such that equation (\ref{eqn:PACAS}) is satisfied. Equation (\ref{eqn:PACAS}) can be interpreted as follows:

\begin{itemize}
    \item {\bf Accuracy.} The inner inequality ``$P_{\Delta_{1,x_2,u}}(\cdots) \geq 1-\mu$" asserts that for a sample drawn at random from $\mathcal{R}^{-1}(x_2) \subseteq X_1$ according to $z_{x_2}$, and for the successor $\Delta_1(z_{x_2},u_1)$ of $z_{x_2}$ under $u_1$, the probability of the existence of a successor $x_2' \in \Delta_2(x_2,u_2)$ of $x_2$ under $u_2$ that is related to  $\Delta_1(z_{x_2},u_1)$ is greater than $\ge 1-\mu$. This quantity serves as a probabilistic measure of the \emph{accuracy} of the alternating simulation relation, implying that a given concrete transition is simulated by an abstract transition with 
    probability 
    of at least $1-\mu$.
    \item {\bf Confidence.} 
    $P_{\Delta_1(z_{x_2}^{(1)}, u_1),\dotsc,\Delta_1(z_{x_2}^{(M)}, u_1)}(\cdots)$  in the outer inequality
    is the probability that the samples $\Delta_1(z_{x_2}^{(1)}, u_1),\dotsc,\Delta_1(z_{x_2}^{(M)}, u_1)$ are sufficiently informative about the transitions in order to construct an abstraction of $\ge 1-\mu$ accuracy. Thus, the probability $1-\delta/{n_x n_u}$ may be taken as the \emph{confidence} that a given transition in the abstraction achieves $1-\mu$ accuracy. Equivalently, the quantity $\delta /{n_x n_u}$ upper bounds the probability that a given transition is not $(1-\mu)$-accurate. By a union bound argument, the probability that any one of the $n_x n_u$ transitions is not $(1-\mu)$-accurate is at most $\delta$. Therefore, the probability that all of the transitions in the abstraction achieve a probabilistic accuracy of at least $1-\mu$ is at least $1-\delta$.
\end{itemize}

One can readily see that when $\mu$ and $\delta$ converge to zero, the proposed PAC approximate alternating simulation recover the classical notion of approximate alternating simulation relation in Definition~\ref{Def:altsimu}.
In the following result we use the concept of PAC alternating simulation  to refine a controller satisfying the specification from one system to the other.

\begin{theorem}
\label{thm:refinement}
Let $S_1=(X_1,X_1^0,U_1,\Delta_1,Y_1,H_1)$ be a deterministic transition system and $S_2=(X_2,X_2^0,U_2,\Delta_2,Y_2,H_2)$ be a finite transition system such that $S_2\preccurlyeq^{PAC}_{(\varepsilon,\delta,\mu)} S_1$, with a PAC approximate alternating simulation relation $\mathcal{R}$,
for probability distributions $z_{x_2},\ x_2\in X_2$, and constants $\varepsilon \geq 0$, and $\delta,\mu \in (0,1)$. Consider a specification $\phi \subseteq Y_2^{\leq N}$ on the output behaviour of the transition system $S_2$.
If there exists a control policy
$(X^0_{\mathcal{C}_2},\mathcal{C}_2)$ such that the closed loop behaviour $\mathcal{B}_{S_2}(X^0_{\mathcal{C}_2},\mathcal{C}_2)$ of the system $S_2$ under $(X^0_{\mathcal{C}_2},\mathcal{C}_2)$ satisfies the specification $\phi$, then there exists a control policy $(X^0_{\mathcal{C}_1},\mathcal{C}_1)$
such that any closed-loop behaviour $(x_{1,0},u_{1,0}),(x_{1,1},u_{1,1}),\dotsc,(x_{1,m},u_{1,m})$, with length $m \leq N$, of $S_1$ under $(X^0_{\mathcal{C}_1},\mathcal{C}_1)$ satisfies
\begin{equation}
\label{eq:behavioural_pac}
\begin{split}
    P_S(&P_u((H_1(x_{1,0}),\dotsc,H_1(x_{1,m})) \\ &\in \mathcal{B}_{\varepsilon}(\phi)) \geq (1-\mu)^m) \geq 1-\delta,
\end{split}
\end{equation}
where $P_u$ denotes the product measure of $P_{\Delta_{1,x_{2,i},u_{2,i}}},\ i=1,\dotsc,m$, $x_{2,i}=\mathcal{R}(x_{1,i})$, $i=1,\dotsc,m$, $P_S$ denotes the product measure of $P_{\Delta_1(z_{x_2}^{(1)},u_2),\dotsc,\Delta_1(z_{x_2}^{(M)},u_2)}$ for all $x_2\in X_2$, $u_2\in U_2$, and $H_1$ is the output map of the transition system $S_1$.
\end{theorem}

\medskip

\begin{proof}
Let the set of initial condition $X^0_{\mathcal{C}_1}=\mathcal{R}(X^0_{\mathcal{C}_2})$. Using the fact that $X^0_{\mathcal{C}_2}$ is non-empty, one has from (i) in Definition~\ref{Def:PACaltsimu} that $X^0_{\mathcal{C}_1}$ is non-empty. By construction of the set $X^0_{\mathcal{C}_1}$, we have for any $x_{1,0} \in X^0_{\mathcal{C}_1}$ the existence of $x_{2,0} \in X^0_{\mathcal{C}_2}$ such that $(x_{1,0},x_{2,0})\in \mathcal{R}$. By iterating $m$ times condition (iii) in Definition~\ref{Def:PACaltsimu}, the inner probability in~\eqref{eq:behavioural_pac} can be factored as
\begin{align}
\label{eq:pu_decomposition}
    & \quad P_u\Big(H_1(x_{1,0}),\dotsc,H_1(x_{1,m}))\in \mathcal{B}_{\varepsilon}(\phi)\Big)\\
    &\geq P_u\bigg( 
    \bigtimes_{i=1}^m
    \{x_{1,i} = \Delta_1(x_{1,i-1}, u_{1,i-1}):\nonumber\\
    &\quad\exists x_{2,i}\in\Delta_2(x_{2,i-1}, u_{2,i-1})
        \text{ s.t. } (x_{1,i},x_{2,i}) \in \mathcal{R}
    \}
    \bigg)\nonumber\\
    \label{eq:pu_decomposition_product}
    & = \prod_{i=1}^m
    P_{\Delta_{1,x_{2,i},u_{2,i}}}
    \bigg(
        x_{1,i} = \Delta_1(x_{1,i-1}, u_{1,i-1}):\\
        &\quad \exists x_{2,i}\in\Delta_2(x_{2,i-1}, u_{2,i-1})
        \text{ s.t. } (x_{1,i},x_{2,i}) \in \mathcal{R} 
    \bigg).\nonumber
\end{align}
where the first inequality follows from (ii) in Definition~\ref{Def:PACaltsimu} and the fact that $H_2(x_{2,0}),\dotsc,H_2(x_{2,m}) \in \phi$, and the first equality follows from the construction of the product measure $P_u$. 

Next, we show that the factors in the product~\eqref{eq:pu_decomposition_product} can with probability $1-\delta$, all be lower-bounded by $1-\mu$. Let $\mathcal{H}$ denote a collection of $M$ samples for each discrete state and input, that is $\mathcal{H}=\{(\Delta_1(z_{x_2}^{1}, u_2),\dotsc,\Delta_1(z_{x_2}^{M}, u_2)):x_2\in X_2, u\in U_2\}$. Since $X_2$ and $U_2$ are finite, and $\Delta_1$ is a deterministic transition relation, the set $\mathcal{H}$ is a finite set. Let $A_{x_2,u_2}$ denote the set of samples $\mathcal{H}$ such that the inner inequality in (\ref{Def:PACaltsimu}) holds for the pair $(x_2,u_2)$. Then we have
\begin{equation}
    P_S(A_{x_2,u_2}^c)= P_{\Delta_1(z_{x_2}^{(1)}, u_2),
		   \dotsc,
		   \Delta_1(z_{x_2}^{(M)},u_2)}
		   (A_{x_2,u_2}^c)=\delta/{n_x n_u},
\end{equation}
where $A^c$ denotes the set-theoretic complement of the set $A$. The first equality holds because the event that the inner inequality in (\ref{Def:PACaltsimu}) holds for a fixed $(x_2,u_2)$ depends only on the samples $\Delta_1(z_{x_2}^{(1)},u_2),\dotsc,\Delta_1(z_{x_2}^{(M)},u_2)$.

The event that the inner inequality in (\ref{Def:PACaltsimu}) fails to hold for at least one $(x_2,u_2)$ pair is the union of the $n_x n_u$ events $\bigcup_{x_2\in X_2, u_2\in U_2}A_{x_2,u_2}^c$, whose probability can be bounded by the union bound:
\begin{equation}
    P_S\big(\bigcup_{x_2\in X_2, u_2\in U_2}A_{x_2,u_2}^c\big)
    \le
    \sum_{x_2\in X_2, u_2\in U_2} P_S(A_{x_2,u_2}^c)=\delta.
\end{equation}
Therefore, the event that the inner inequality in (\ref{Def:PACaltsimu}) holds for all $(x_2,u_2)$ is at least $1-\delta$, which we can also express as $P_S( \mathcal{H}\notin \bigcup_{x_2\in X_2, u_2\in U_2}A_{x_2,u_2}^c) \ge 1-\delta$.

Now, suppose $\mathcal{H} \notin \bigcup_{x_2\in X_2, u_2\in U_2}A_{x_2,u_2}^c$: then we may apply the inner inequality in (\ref{Def:PACaltsimu}) to each factor of the product in~\eqref{eq:pu_decomposition_product}, yielding
\begin{equation}
P_u\Big(H_1(x_{1,0}),\dotsc,H_1(x_{1,m}))\in \mathcal{B}_{\varepsilon}(\phi)\Big)\le(1-\mu)^m.
\end{equation}
Since $P_S( \mathcal{H}\notin \bigcup_{x_2\in X_2, u_2\in U_2}A_{x_2,u_2}^c) \ge 1-\delta$, we have
\begin{equation}
\begin{split}
    P_S(&P_u((H_1(x_{1,0}),\dotsc,H_1(x_{1,m})) \\ &\in \mathcal{B}_{\varepsilon}(\phi)) \geq (1-\mu)^m) \geq 1-\delta.
\end{split}
\end{equation}

\end{proof}

\medskip
The result of Theorem~\ref{thm:refinement} can be interpreted as follows: if a transition system $S_2$ is related to a transition system $S_1$ by a $(\varepsilon,\delta,\mu)$-PAC approximate alternating simulation relation and if one can synthesize a controller for $S_2$ to achieve a specification $\phi$, then we can refine the controller for $S_2$ into a controller for $S_1$ ensuring, with a confidence $1-\delta$, that the system $S_1$ will satisfy an approximate version of the specification $\phi$ (given by $\mathcal{B}_{\varepsilon}(\phi)$).

\subsection{PAC symbolic abstraction}

Given the concrete system $S=(X,X^0,U,\Delta,Y,H)$ 
with an unknown transition function $\Delta$, a precision $\varepsilon \geq 0$, a confidence parameter $\delta \in (0,1)$, and an accuracy $\mu \in(0,1)$, in this section we propose a randomized algorithm to construct a symbolic abstraction
$S_d=(X_d,X_d^0,U_d,\Delta_d,Y_d,H_d)$ that is related to concrete system $S$ by a $(\epsilon,\mu,\delta)$-PAC alternating simulation relation. This algorithm is a data-driven generalization of the classical abstraction algorithm described in Section~\ref{subsec:symbolic_abstraction}.

The algorithm is made up of two steps:
\begin{enumerate}
    \item We construct the discrete set of states $X_d$, initial states $X_d^0$, inputs $U_d$, outputs $Y_d$ and output map $H_d$ following the same approach in step 1 of Section~\ref{subsec:symbolic_abstraction}\label{ln:discretization}. 
    
    \item To construct the discrete transition relation $\Delta_d$, we first select a sample size $M \in \N_{>0}$ according to (\ref{eqn:samplesize}). Then for a discrete state $q=(\underline{x}^q,\overline{x}^q] \in X_d$ we select a distribution $p_q$ such that for $x \in X$, $p_q(x) > 0 \Longleftrightarrow x \in q=(\underline{x}^q,\overline{x}^q]$. We then select a discrete input $u \in U_d$, draw $M$ samples $x^{(1)},\dotsc,x^{(M)}$ from the set $(\underline{x}^q,\overline{x}^q]$ according to the distribution $p_q$, and compute the successors $\Delta(x^{(1)},u),\dotsc,\Delta(x^{(M)},u)$. The discrete transition relation is then defined as follows: for $q,q' \in X_d$, $u \in U_d$, $q' \in \Delta_d(q,u)$ if and only if there exists $i\in\{1,\dotsc,M\}$ such that $\Delta(x^{(i)},u)\in q'$.
\end{enumerate}

\begin{figure}[!t]
	\begin{center}
		\includegraphics[width=1\columnwidth]{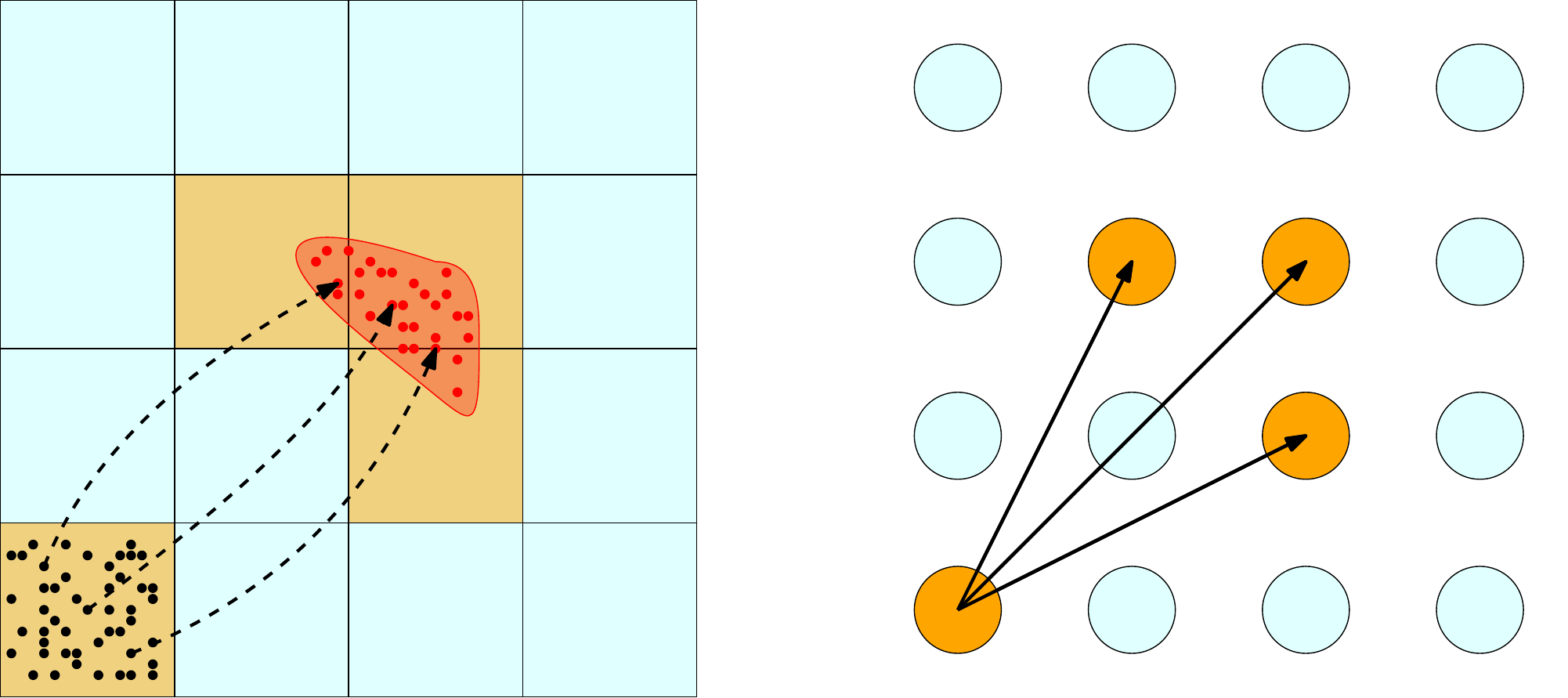}\\
	\end{center}
	\caption{Illustration of the construction of the symbolic abstraction $S_d$. Starting from a discrete state $q=(\underline{x}^q,\overline{x}^q]$ represented by the bottom left orange rectangle, we draw $M$ samples $x^{(1)},\dotsc,x^{(M)}$ from $q$ according to the distribution $p_q$ and compute the continuous successors $\Delta(x^{(1)},u),\dotsc,\Delta(x^{(M)},u)$ under a given control input $u$. The set of discrete successors of $q$ under the control input $u$ is the set of discrete states that include the continuous successors $\Delta(x^{(1)},u),\dotsc,\Delta(x^{(M)},u)$. In this example, we can see that all the continuous successors are included in $3$ discrete states. Hence, we have $3$ successors of the discrete state $q$ under the input $u$.}
	\label{fig:abstraction}	
\end{figure}

The construction above is also summarized in pseudo-code format in Algorithm~\ref{alg:data_driven_abstraction} and illustrated in Figure~\ref{fig:abstraction}.

We have the following result relating the concrete transition system $S$ to its symbolic abstraction $S_d$.

\begin{theorem}
\label{thm:abstraction_is_pac}
Given a concrete system $S=(X,X^0,U,\Delta,Y,H)$, 
a precision $\varepsilon \geq 0$, a confidence parameter $\delta \in (0,1)$, and an accuracy $\mu \in(0,1)$. Let $S_d$ be a symbolic abstraction of the system $S$ constructed according to Algorithm~\ref{alg:data_driven_abstraction}. Then 
$S_d
\preccurlyeq^{PAC}_{(\varepsilon,\delta,\mu)}
S
$.
\end{theorem}

\medskip

\begin{algorithm}[htbp]
    \SetAlgoLined
    \caption{Randomized algorithm to construct an $(\epsilon,\mu,\delta)$-PAC symbolic abstraction}
    \label{alg:data_driven_abstraction}
    \KwIn{Concrete system $S=(X,X^0,U,\Delta,Y,H)$; parameters $\epsilon>0$, $\mu,\delta\in(0,1)$.}
    \KwOut{Abstract system $S_d=(X_d,X_d^0,U_d,\Delta_d,Y_d,H_d)$ satisfying $S_d \preccurlyeq^{PAC}_{(\varepsilon,\delta,\mu)} \Sigma $.}
    Construct $U_d$, $X_d$, $X^0_d$ and $H_d$, and $Y_d$ according to the procedure in step 1 of Section~\ref{subsec:symbolic_abstraction}\label{ln:discretization} with a state-space discretization parameter $\varepsilon$, and with $\card(X_d)=n_x$ and $\card(U_d)=n_u$\;
    Select a sample size
        $M \ge \frac{1}{\mu}\left(n_x\log 2 + \log\frac{n_x n_u}{\delta}\right)$\label{ln:choose_sample_size}\;
    \ForAll{$q \in X_d$} {
        Select a distribution $p_q$ such that for $x \in X$ $p_q(x) > 0 \Longleftrightarrow x\in q$\label{ln:select_distribution} \;
        \ForAll{$u\in U_d$}{
        Draw $M$ samples $x^{(1)},\dotsc,x^{(M)}$ from $q$ according to the distribution $p_q$, and compute the successors $\Delta(x^{(1)},u),\dotsc,\Delta(x^{(M)},u)$\;
        Define $\Delta_d(q,u) = \{q'\in X_d : \exists i\in\{1,\dotsc,M\}\text{ s.t } \Delta(x^{(i)},u)\in q'\}$\label{ln:construct_transition}\;
        }
    }
\end{algorithm}	
 
\begin{proof}
Consider the relation $\mathcal{R} \subseteq X \times X_d$ defined for $(x,q) \in X\times X_d$ as $(x,q) \in \mathcal{R}$ if and only if $x \in q$. Let us show that $\mathcal{R}$ is an $(\varepsilon,\delta,\mu)$-PAC approximate alternating simulation relation from $S_d$ to $S$. First, we have that, for any $q \in X_d$, there exists an $x \in X$ such that $x \in q$: hence condition (i) of Definition~\ref{Def:PACaltsimu} is directly satisfied. Now for $(x,q) \in \mathcal{R}$ with $q=(\underline{x}^q,\overline{x}^q]$, by considering $\mathbf{d}$ to be the Euclidean distance on $X \subseteq \R^n$, we have that $\mathbf{d}(H(x),H_d(q))=\mathbf{d}(x,\frac{\underline{x}^q+\overline{x}^q}{2}) \leq \varepsilon$, where the inequality follows from the fact that we are using $\varepsilon$ as a state-space discretization parameter. Hence, condition (ii) of Definition~\ref{Def:PACaltsimu} is satisfied. Let us now show that condition (iii) in Definition~\ref{Def:PACaltsimu} is satisfied. Consider $(x,q)\in \mathcal R$, $u_d \in U^a(q)$ and choose $u=u_d \in U^a(x)$. 

To prove that equation (\ref{eqn:PACAS}) is satisfied, we show that (\ref{eqn:PACAS}) is equivalent to a PAC bound of the form asserted by Proposition~\ref{prop:finite_concept_pac} for an empirical risk minimization problem that is solved by the construction of $\Delta_d(q,u_d)$ in step~\ref{ln:construct_transition} of Algorithm~\ref{alg:data_driven_abstraction}. First, we note that
\begin{align}
\label{eq:behavioralpac_to_erm_1}
    &P_{\Delta_{q,u}}\Big( x'=  \Delta(x,u):\; \nonumber\\&\quad \exists q'\in  \Delta_d(q,u) \text{ s.t } (x',q')\in \mathcal R\Big)\\
    &=P_{\Delta_{q,u}}\Big(\bigcup_{q'\in\Delta_d(q,u)} \mathcal{R} ^{-1}(q')\Big),
    \label{eq:behavioralpac_to_erm_3}
\end{align}
by the definition of $\mathcal{R}^{-1}$.
Now, consider the probability in~\eqref{eq:behavioralpac_to_erm_3}: this is the complement of the risk function corresponding to an empirical risk minimization problem whose concept class is the set of $2^{n_x}$ unions of the form $\bigcup_{q' \in J} \mathcal{R}^{-1}(q')$ indexed by $J\subseteq X_d$, whose loss function is the set membership loss, and whose data are the successor samples $\Delta(x^{(1)},u),\dotsc,\Delta(x^{(M)},u)$. For this problem, the empirical risk for the concept corresponding to the discrete states $J\subseteq X_d$ is
\begin{equation}
    \hat{r}(J) = \frac{1}{N}\sum_{i=1}^M
    \ind{\Delta(x^{(i)},u) \notin \bigcup_{q' \in J} \mathcal{R}^{-1}(q')}.
\end{equation}
By construction, the set $\bigcup_{q'\in \Delta_d(q,u)} \mathcal{R}^{-1}(q')$ satisfies $\hat{r}(\Delta_d(q,u))=0$, since $q'\in \Delta_d(q,u)$ only if $\Delta(x^{(i)},u)\in \mathcal{R}^{-1}(q')$ for some $i\in\{1,\dotsc,M\}$.

To attain the PAC bound, we now apply Proposition~\ref{prop:finite_concept_pac} to the empirical risk minimization problem described above. We must first check that the sample size is sufficient: since there are $2^{n_x}$ concepts, and we want to ensure accuracy $\mu$ and confidence $\delta/{n_x n_u}$, we must have
\begin{equation}
\label{eqn:samplesize}
M
\ge\frac{1}{\mu}\log\frac{2^{n_x}}{\delta/n_x n_u}
=\frac{1}{\mu}\left(n_x\log 2 + \log\frac{n_x n_u}{\delta}\right),
\end{equation}
which is precisely the sample size requirement stipulated in Algorithm~\ref{alg:data_driven_abstraction}. Therefore, by Proposition~\ref{prop:finite_concept_pac} and the fact that $\hat{r}(\Delta_d(q,u))=0$, we have
\begin{equation}
\begin{aligned}
\label{eq:algorithm_erm_pac}
    &P_{\Delta_{q^{(1)},u},\dotsc,\Delta_{q^{(M)},u}} \Big( \\
    &\quad 1 - 
    P_{\Delta_{q,u}}\Big(\bigcup_{q'\in\Delta_d(q,u)} \mathcal{R} ^{-1}(q')\Big) \le \mu \Big) \ge 1-\delta/{n_x n_2}.
\end{aligned}
\end{equation}
This inequality is equivalent to
\begin{equation}
\begin{aligned}
\label{eq:algorithm_erm_pac}
    &P_{\Delta_{q^{(1)},u},\dotsc,\Delta_{q^{(M)},u}} \Big(\\ &\quad
    P_{\Delta_{q,u}}\Big(\bigcup_{q'\in\Delta_d(q,u)} \mathcal{R} ^{-1}(q')\Big) \ge 1 - \mu \Big) \ge 1-\delta/{n_x n_u},
\end{aligned}
\end{equation}
which in turn is equivalent to~\eqref{eq:behavioural_pac} by the reasoning of Eqs.~\eqref{eq:behavioralpac_to_erm_1} through~\eqref{eq:behavioralpac_to_erm_3}.

\end{proof}

\medskip

Note that any precision $\varepsilon>0$ of the PAC alternating simulation relation can achieved by choosing an appropriate state-space discretization parameter. Moreover, for a given precision $\varepsilon>0$, a confidence parameter $\delta \in (0,1)$, and an accuracy $\mu \in(0,1)$, there always exists $M \in \mathbb{N}$ sufficiently large such that equation (\ref{eqn:samplesize}) holds, which reflects the fact that any confidence parameter $\delta$ and accuracy $\mu$ can be achieved by an appropriate choice of the sample size $M$.

\section{Numerical example}
\label{sec:4}

In this section, we demonstrate the practicality of our approach on a marine vessel control problem~\cite{meyer2021abstraction}. In the following, the numerical implementations has been done in MATLAB, Processor 2.6 GHz 6-Core Intel Core i7, Memory 16 GB 2667 MHz DDR4.

The dynamics of the kinematic model of the marine vessel is given by:

\begin{align*}
  \dot{x}_1 &= u_1\cos(x_3)-u_2\sin(x_3)+d_1 \\
  \dot{x}_2 &= u_2\sin(x_3)+u_2\cos(x_3)+d_2 \\
  \dot{x}_3 &=u_3+d_3
\end{align*}

The state of model consists of the planar position $(x_1,x_2)$ and
the heading $x_3$ of the marine vessel. The control inputs $u \in \R^3$ are the surge velocity, sway velocity, and yaw rate.

We consider a state space defined as the interval $X=[0,10]\times [0,6.5]\times [-\pi,\pi]$ with a target subset $X_T=[7,10]\times [0,6.5]\times [\pi/3,2\pi/3]$  and two static obstacles $X_{O_1}=[2,2.5]\times [0,3]\times [-\pi,\pi]$ and $X_{O_2}=[5,5.5]\times [3.5,6.5]\times [-\pi,\pi]$. The control objective is to reach the target set $X_T$ in less than $2$ minutes while remaining in the safe region $X \setminus (X_{O_1}\cup X_{O_2})$. The input constraints are given by $U=[0,0.18]\times [-0.05,0.05]\times [-0.1,0.1]$. 

\begin{figure}[!t]
	\begin{center}
		\includegraphics[width=1\columnwidth]{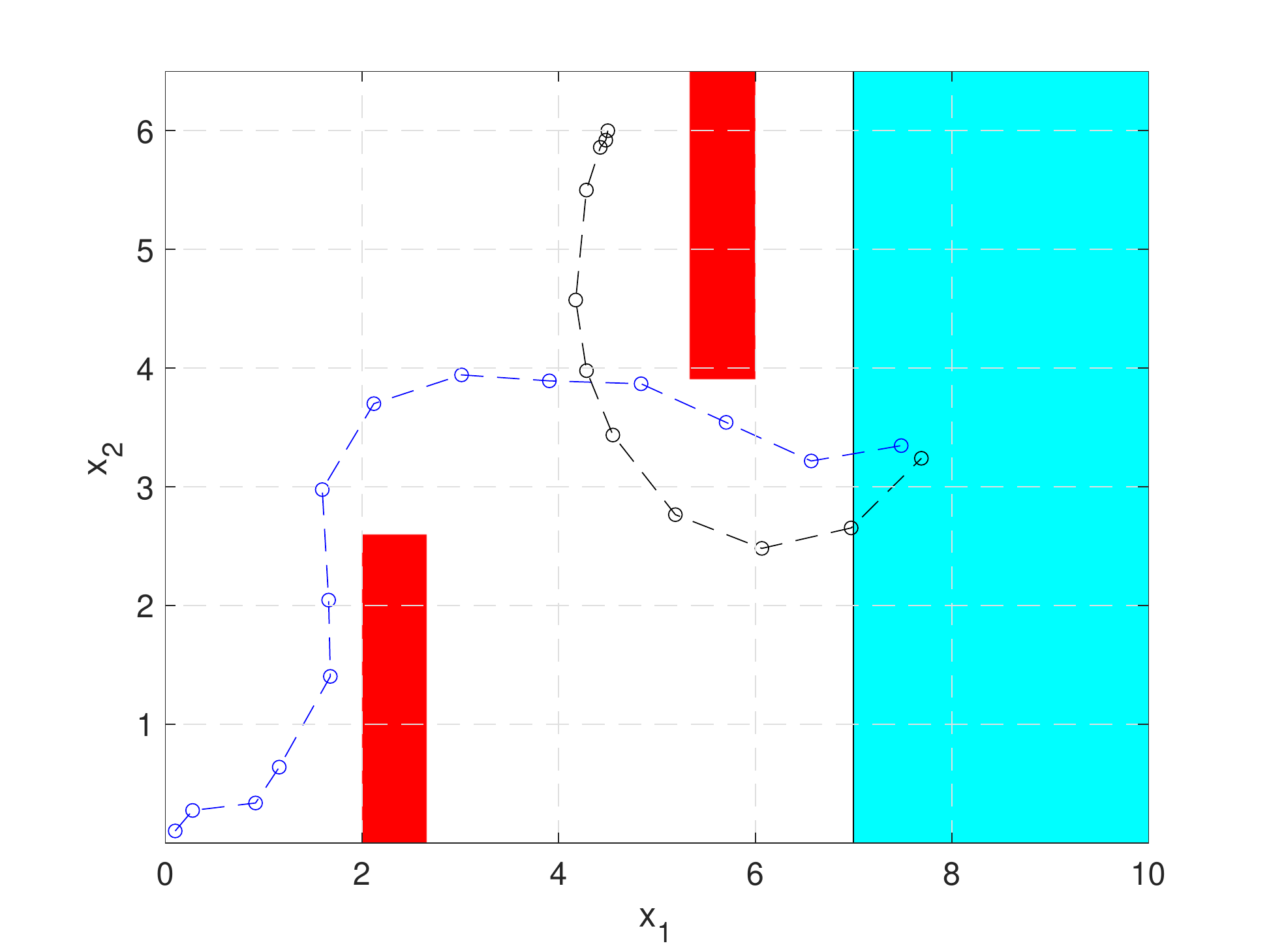}\\
	\end{center}
	\caption{Illustration of two controlled trajectories of the marine vessel in the $x_1-x_2$ plane, the target set (light blue region) and the obstacles (red regions). }
	\label{fig:hybrid}	
\end{figure}

From the continuous-time model described above, we generate a discrete-time model using the sampling period $\tau=5 s$. Then, we use the symbolic approach presented in Algorithm~\ref{alg:data_driven_abstraction} to construct a symbolic model of the marine vessel model. For the symbolic model, a controller is synthesized to achieve the time bounded reach avoid specification by combining the safety and reachability games (See~\cite[Sections 6.2 and 6.3]{tabuada2009verification}). The abstract controller is then refined into a concrete controller ensuring the satisfaction of the control objective. The parameter of the construction of the abstraction are the following. The precision $\varepsilon=1$, the number of discrete states is $n_x=3375$ (corresponding to $15$ interval per dimension) and the number of discrete inputs is $n_u=64$ (corresponding to $4$ inputs per dimension). For the construction of the transition relation, we choose a confidence parameter $\delta=10^{-6}$, and an accuracy $\mu=0.1$. This choice of $\mu$ and $\delta$ ensures, by Theorem~\ref{thm:abstraction_is_pac}, that the event that any one of the estimated transitions achieves less than $90\%$ probabilistic accuracy is a ``one in a million'' event. According to step~\ref{ln:choose_sample_size} in Algorithm~\ref{alg:data_driven_abstraction}, the number of samples should be greater than $M=23655$. The computation time for generating the symbolic model took $18$ hours and the controller synthesis took $4$ minutes. The controllable space for the reach–avoid specification covers more than $92\%$ of the
state space. Figure~\ref{fig:hybrid} provides two examples of trajectories of the controlled system. For the trajectory in blue, the system is initialized at $x^0=[0.1;0.1;-\frac{\pi}{3}]$ and for the trajectory in black, the system is initialized at $x^0=[4.5;6;0]$. One can readily see that the controlled trajectories avoid the obstacles (in red) and eventually reaches the target set corresponding to the 2D docking area (in light blue).

\section{Conclusion}
In this paper, a new data-driven approach for computing guaranteed finite abstractions has been proposed. The proposed approach does not require closed-form dynamics, but instead only the ability to evaluate successors of individual points for given inputs. A new PAC behavioural relationship has been proposed, with an algorithmic procedure to construct symbolic abstractions. Statistical guarantees are made on the constructed symbolic model in terms of accuracy and confidence, where any degree of precision can be achieved by appropriately choosing the number of required data. An illustrative example shows the merits of the proposed approach.

Several directions will be explored in future works. A first objective is to generalize the results to deal with systems under disturbances.
Another objective which is critical for practical application is to reduce the size of the sample bound of Algorithm~\ref{alg:data_driven_abstraction}.
The result of Theorem~\ref{thm:abstraction_is_pac} allows for the possibility that the discrete transition could be any union of discrete cells: with some additional system information, the number of possible transitions, and thereby the number of samples needed to ensure a given statistical accuracy and confidence could be significantly reduced. 
Additionally, Algorithm~\ref{alg:data_driven_abstraction} possesses a parallel structure which was not fully leveraged in this paper. Specifically, steps 3 through 8 of Algorithm~\ref{alg:data_driven_abstraction} may be executed in parallel for each $q\in X_d$, since construction of the discrete transition $\Delta_d(q,u)$ for the state $q$ does not depend on the transitions for any other states. Since the number of discrete states is typically very large, this state-level parallelism is sufficient to fully utilize the parallel capabilities of many high-performance computing systems, such as those available in the AWS EC2 platform.
Another advantage of the statistical approach is that it does not rely on the methods of reachability analysis, which are typically restricted to some specific class of sets~\cite{althoff2020set} which can reduce the fidelity of the discrete transitions. Effectively, the fidelity of the statistical approach is limited only by the fidelity of the discretization. Finally, in order to deal with large scale interconnected systems, a compositional data-driven approach can be developed along the same lines of~\cite{saoud2021compositional}.

\bibliographystyle{IEEEtran}
\bibliography{IEEEabrv,bibfile}

\end{document}